\documentclass[10pt,letterpaper]{article}
\usepackage[top=0.85in,left=2.75in,footskip=0.75in,marginparwidth=2in]{geometry}

\usepackage[utf8]{inputenc}

\usepackage{cite}

\usepackage{amsmath}

\usepackage{nameref,hyperref}

\usepackage[right]{lineno}

\usepackage{microtype}
\DisableLigatures[f]{encoding = *, family = * }

\raggedright
\usepackage{changepage}

\usepackage[aboveskip=1pt,labelfont=bf,labelsep=period,singlelinecheck=off]{caption}

\makeatletter
\renewcommand{\@biblabel}[1]{\quad#1.}
\makeatother

\usepackage{lastpage,fancyhdr,graphicx}
\usepackage{epstopdf}
\fancyheadoffset[L]{2.25in}
\fancyfootoffset[L]{2.25in}

\usepackage{color}

\definecolor{Gray}{gray}{.25}

\usepackage{graphicx}

\usepackage{sidecap}

\usepackage{wrapfig}
\usepackage[pscoord]{eso-pic}
\usepackage[fulladjust]{marginnote}
\reversemarginpar

\begin{document}
\vspace*{0.35in}

\begin{flushleft}
{\Large
\textbf\newline{\ Hashing it Out: Predicting Unhealthy Conversations on Twitter}
}
\newline
\\
Steven Leung\textsuperscript{1},
Filippos Papapolyzos\textsuperscript{2}
\\
\bigskip
\bf{1} UC Berkeley
\\
\bf{2} UC Berkeley
\\
\bigskip
stevendleung@berkeley.edu

\end{flushleft}

\section*{Abstract}
Personal attacks in the context of social media conversations often lead to fast-paced derailment, leading to even more harmful exchanges being made. State-of-the-art systems for the detection of such conversational derailment often make use of Deep Learning approaches for prediction purposes. In this paper, we show that an Attention-based BERT architecture, pre-trained on a large Twitter corpus and fine-tuned on our task, is efficient and effective in making such predictions. This model shows clear advantages in performance to the existing LSTM model we use as a baseline. Additionally, we show that this impressive performance can be attained through fine-tuning on a relatively small, novel dataset, particularly after mitigating overfitting issues through synthetic oversampling techniques. By introducing the first transformer based model for forecasting conversational events on Twitter, this work lays the foundation for a practical tool to encourage better interactions on one of the world's most ubiquitous social media platforms.

\section{Introduction}

Social media has become one of the primary stages where peer-to-peer discussions take place, spanning anything from everyday local matters to high level philosophical topics. A very frequent phenomenon that is noticed in such conversations is their rapid deterioration upon the presence of a personal attack, often leading to personal threats and insults which heavily undermine the level of discussion. In order to battle this issue, Social Media companies, such as Twitter and Facebook, have come up with systems that make use of human moderators who review content flagged by users. In the majority of cases, content that goes against the company’s policy is removed \cite{twitter}.  

The way this moderation tactic is set up presents a series of challenges and limitations, the most obvious being the cost of moderation services. Secondly, there have been cases of unreliable moderation in Twitter, with users receiving temporary bans simply for mentioning some specific word in their tweets. Lastly, it has to be appreciated that  moderation can be quite an emotionally taxing job as content under review will often be very disturbing.

The most basic limitation of this approach, however, is that moderation only happens a posteriori and solely after users choose to report the abusive content. This leaves a significant space in time in which a conversation can escalate further and users might become the victims or perpetrators of personal attack and/or verbal abuse. To the best of our knowledge, no effort is done on the part of Twitter in order to prevent such harmful content rather than moderate it. We trust that if Social Media companies chose to assume a preventative strategy to moderation, this would reduce the instances of personal attacks and improve the overall content quality.

There are a series of issues that make detecting potentially harmful language a challenging topic. As mentioned in Twitter’s enforcement policy \cite{twitter}, context is of utmost importance; the use of specific language might have a different interpretation when the preceding conversation is better examined. Twitter’s policy mentions that factors such as whether “the behavior is directed at an individual, group, or protected category of people” and “the severity of the violation” are also considered prior to taking action. 

The aim of this project is to provide a Deep Learning approach to content moderation, by means of an attention-based neural network model that predicts whether a Twitter conversation is likely to deteriorate and lead to a personal attack. We believe that this will not only provide an advantage to moderators but could also potentially be utilized to give warnings to users when they are about to tweet something which may lead to escalation.

\section{Related Work}

A lot of our inspiration for this project was drawn from the 2018 paper Conversations Gone Awry: Detecting Early Signs of Conversational Failure (Zhang et al., 2018) \cite{zhang-etal-2018-conversations}. Specifically, in this paper the authors try to identify predictors of conversational derailment and make predictions using a variety of strategies. An interesting insight from the paper is that conversations containing an attacker-initiated direct question are most likely to derail. The authors also claim that human-based predictions of conversational derailment are at an accuracy level of 72\%. 

Past work has also been done on forecasting hostility on a dataset of Instagram comments (Liu et al., 2018) \cite{liu2018forecasting}, analyzing features of personal attack in Wikipedia comments (Wulczyn et al., 2017) \cite{wulczyn2017ex} and the use of attention models to improve the performance of RNN and CNN approaches to content moderation (Pavlopoulos et al., 2017) \cite{pavlopoulos-etal-2017-deeper}.

In the 2019 paper titled Trouble on the Horizon: Forecasting the Derailment of Online Conversations as they Develop, Chang et al. \cite{chang2019trouble}, the authors use a Recurrent Neural Network (RNN) based approach to predict conversational derailment on a dataset of Wikipedia talk page discussions developed as part of the aforementioned 2018 paper, and a dataset of conversations in the ChangeMyView subreddit. CRAFT, their model architecture, yields an accuracy score of 66.5\% which, to the best of our knowledge, is the highest accuracy score achieved on this specific task. 

Despite the success of the RNN-based CRAFT model, we choose to explore a pre-trained self-attention model for our conversational task for several reasons. First, work by Vaswani et al. \cite{DBLP:journals/corr/VaswaniSPUJGKP17}, and others have shown performance advantages of transformers vs RNNs. Namely, RNNs have displayed challenges with long range dependencies and word sense disambiguation relative to transformers \cite{tang2018selfattention}. We deemed that using an Attention model could allow us to curb the sequential architecture of RNNs, allowing the model to make use of previous context without information loss. This is grounded in the idea that previous tweets in a conversation may have an equally strong effect on the probability of derailment of a subsequent tweet. In addition, transfer learning allows us to take advantage of pre-trained models, such as BERTweet, which is trained on millions of examples in the unique language of Twitter, thus improving the overall performance of our approach. Finally, due to the linear scaling self-attention models as opposed to the quadratic scaling of RNNs, our model is more scalable to the vast conversational world of Twitter.

\section{Dataset}
\label{sec:length}

For this project we have compiled our own dataset, consisting of 5656 tweets tagged on one binary dimension of personal attack. Tweets are labelled with conversation ID's, vital for our task of using prior tweets to predict the outcome of a subsequent tweet. We acquired our data through the Twitter API. In order to have a higher probability of finding positive examples of personal attack, we selected conversations from a collection of controversial topics such as Universal Basic Income, abortion, and immigration.

In tagging each tweet as either containing a personal attack or not, we used the Wiktionary definition of a personal attack: "an abusive remark on or relating to somebody's person instead of providing evidence when examining another person's claims or comments". We used three methods in tagging. First, we made use of the participant recruitment platform Mechanical Turk. Secondly, we were fortunate enough to find a dataset on Zenodo titled “Hate speech and personal attack dataset in English social media” \cite{polychronis_charitidis_2019_3520152} that contained tweet level tags. We were able to match these tweets to the rest of the conversation through the Twitter API. Finally, we tagged the remainder of tweets ourselves.

From the 5656 tweets, we isolated 1177 positive examples, leaving 4479 negative examples which translated to a heavy class imbalance. Our budget and project time frame were significant limitations to the quality and scale of our data collection, forcing us to seek alternative methods of reducing the effect of positive example under-representation, which came at their own costs. Specifically, we made use of oversampling for positive examples, which came at the expense of overfitting.

\subsection{Synthetic Oversampling}

A method we made use of to reduce class imbalance in our dataset was a technique called synthetic oversampling. Specifically, we deemed that by creating artificial context examples for the positive class and supplying them to our model in the training phase we could reduce the effect of overfitting while also providing a larger quantity of plausible training examples. This technique works by replacing words in the original context with similar words to create new contexts and resupplying them to the model as separate training examples. 

Our inspiration was drawn from the 2016 paper A Study of Synthetic Oversampling for Twitter Imbalanced Sentiment Analysis (Ah-Pine \& Soriano, 2016) \cite{ahpine:hal-01504684} in which the authors present a general algorithm for creating synthetic samples as follows: a random tweet x is drawn from a distribution P, its nearest neighbors NN(x) are determined, a neighbor x’ is chosen “according to a probability distribution over NN(x)” and a synthetic sample is created in the form $y = x + a(x’ - x)$, where $a ~ unif(0,1)$. A popular version of this algorithm is known as SMOTE (Synthetic Minority Oversampling TEchnique) (Chawla, 2002) \cite{Chawla_2002}, which assumes a uniform distribution over P. The application of this algorithm in the context of words requires a vector space mapping which, in our case, was achieved using the GloVe Twitter Embeddings.

Specifically, the original context is first broken down into a series of tokens which are tagged by part-of-speech and are filtered for stopwords, which are not further processed. Using the GloVe Twitter embeddings, a dictionary of k-nearest neighbors is computed for each token using the Euclidean distance metric. We randomly pick one of the 3 closest neighbors with non-uniform probability, giving larger probability to the closest neighbor. We then randomly choose the filtered tokens to be replaced in the original context with probability P = 0.2 and repeat this process n times to produce a list of similar but non-identical contexts. Using this process we developed 355 synthetic positive examples.

\section{Transformer Model}

Our general model for forecasting conversational events is the BERT-base BERTweet transformer model, fine-tuned on our conversational Twitter data.

\subsection{Conversational Structure}

\paragraph{Conversations}
For the purposes of our model, a conversation is made up of an initial tweet (what we call a top-level tweet) and all subsequent tweets following that tweet that are in direct reply to one another. While Twitter does allow for the branching of conversations (at any point in a conversation, users can reply directly to the top-level tweet or to a reply, a reply of a reply, etc.) that allows for complex conversational structures, we limit our model to looking at single conversational branches at a time for each input.

Thus, given top-level tweet T1, a conversation consists of a sequence of N tweets $C=\{T_1,...,T_N\}$

\paragraph{Context}
Tweets are the individual components of the conversation, made by a single user. For the sake of forecasting, we are using the two tweets prior to forecast whether the current tweet in question will be a personal attack. The two tweets prior are referred to as the "context". This is the input into our model. 

The two context tweets are concatenated with the $</s>$ token in between to preserve the delineation of tweets. The context is then tokenized using the BERTweet tokenizer, with a normalization feature to handle common conversational elements in Twitter such as username handles, hashtags and urls. The classification token, $[CLS]$, is appended to the front of the input for use in the classification process. Thus, given the tweet in question, $T_N$, the two tweet context, $T_N-1$ and $T_N-2$, is converted to variable M length tokens, yielding the conversational context, 
 $C_N = [CLS], T_{N-1,1},T_{N-1,2},T_{N-1,...},T_{N-1,M} </s>, T_{N-2,1},T_{N-2,2},T_{N-2,...},T_{N-2,M}$

The most recent tweet in the context is placed in the front of the context to avoid truncating the most recent tweet in the case that the context exceeds BERTweet tokenizer's 130 token length limit.

\begin{figure}[htp]
    \centering
    \includegraphics[width=8cm]{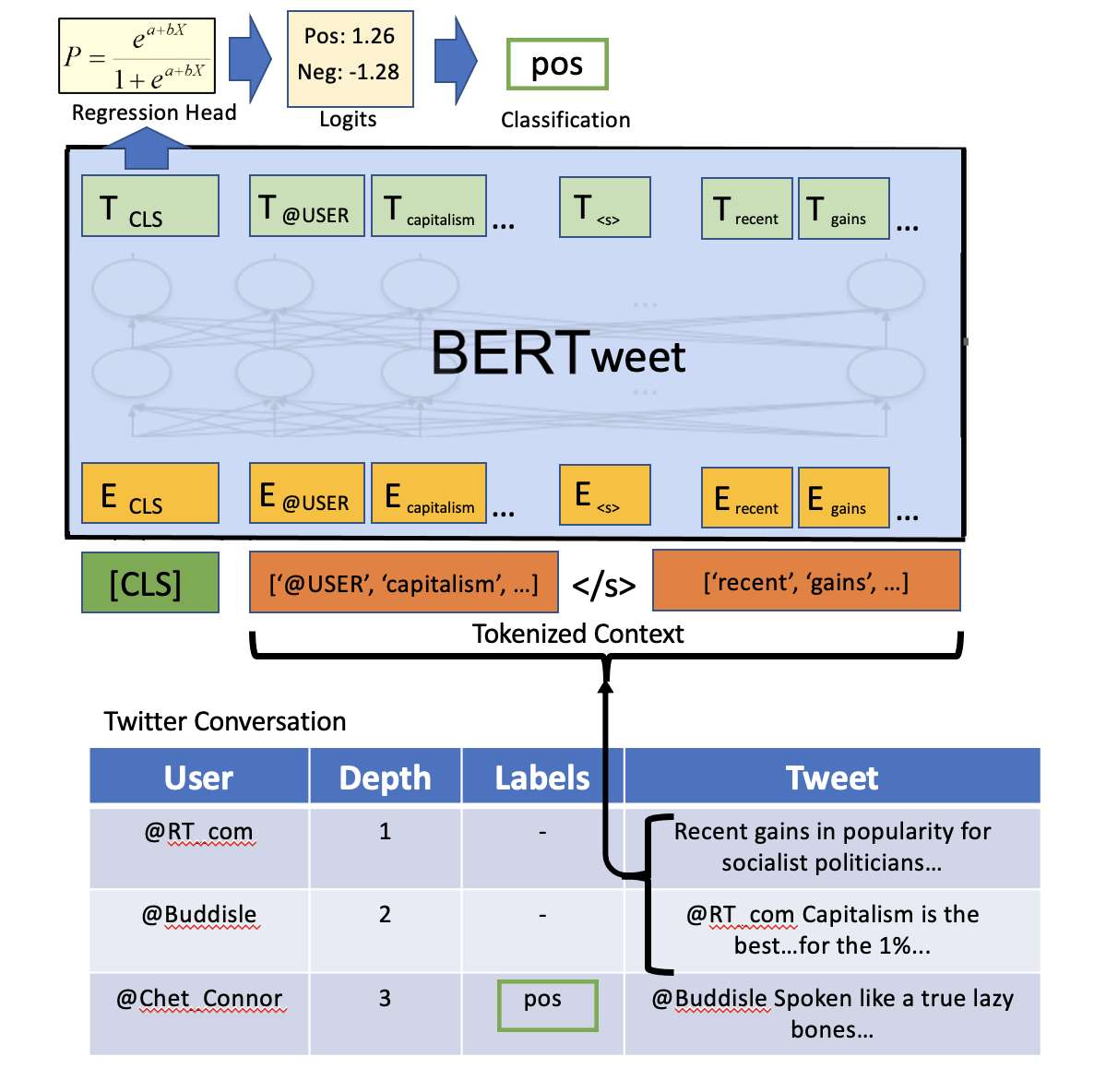}
    \caption{Model architecture. The tokenized two tweet context is fed into pre-trained BERTweet. Forecasting whether the subsequent tweet will be a personal attack or not is treated as a classification problem through a logistic regression head on the CLS token final embedding.}
    \label{fig:galaxy}
\end{figure}

\subsection{Fine-tuning BERTweet}
The pre-trained BERTweet model that we use is a fine-tuned version of the popular BERT model. At a high-level, BERT is a bi-directional encoder model trained to predict masked tokens and whether one sentence directly follows another in the source text \cite{devlin2019bert}. We use the base model configuration consisting of 12 transformer blocks, 12 self-attention heads, and a hidden size of 768. In the BERTweet implementation, BERT base is fine-tuned on 850M tweets for optimal performance on Twitter data in regards to the tasks of part-of-speech tagging, named-entity recognition and text classification \cite{nguyen2020bertweet}. We choose this pre-trained model based on its superior performance to other transformer models such RobertaLarge and XLM-R large, specifically on Twitter data on the tasks mentioned above.

During the fine-tuning process, we update the parameters of BERTweet on the specific task of personal attack forecasting using Twitter conversation context in accordance with the method proposed in Howard et al. \cite{howard2018universal}. We append a classification head on top of BERT using the huggingfaces model configuation Bertforsequenceclassification \cite{wolf2020huggingfaces}. In this implementation, the $[CLS]$ token is used to classify whether the tweet in question is forecasted to be a personal attack or not, in that the final hidden state $h$ of this token is used to classify the sequence.  A simple linear dense layer classification head is added to the top of BERTweet to forecast the probability of a personal attack before passing to a sigmoid transformation for binary classification: \begin{equation} 
\begin{split}
{P}(\text{Attack} \mid \text{$h_{1...768}$})
 = \text{Sigmoid}(W\text{$h_{1...768}$} + b)
\end{split}
\end{equation}
This classification method is discussed in further depth by Sun et al. \cite{sun2020finetune}.

\section{Evaluation and Analysis}

\subsection{Results}

We test and evaluate three models to examine the effect of both the model architecture and synthetic oversampling. We use the CRAFT model as our baseline in the same form as proposed by Chang et al.. CRAFT is trained and evaluated on our Twitter conversation data using standard oversampling on the positive class. We then fine-tune and evaluate BERTweet with standard oversampling to measure the impact of a transformer implementation as opposed to the LSTM implementation proposed in CRAFT. Finally, we fine-tune and evaluate BERTweet on our Twitter conversation data with synthetic oversampling of the positive class.

\begin{table}[ht]
\centering
\begin{tabular}{lrrrrr}
\hline
    \textbf{Model} & \textbf{A} &  \textbf{P} &  \textbf{R} &    \textbf{F1} &  \textbf{AUPR} \\
\hline
    CRAFT &      0.76 &       0.52 &    0.57 &  0.54 &  0.55 \\
    BT &      0.82 &       0.62 &    \textbf{0.76} &  0.68 &  0.76 \\
    BT SOS &      \textbf{0.85} &      \textbf{0.69} &    0.72 &  \textbf{0.70} &  \textbf{0.78} \\
\end{tabular}
\caption{Comparison of performance between CRAFT with random oversampling (CRAFT), BERTweet fine-tuned with random oversampling (BT) and BERTweet fine-tuned with synthetic oversampling (BT SOS). Classification threshold of .5.}
\label{tab:caption}
\end{table}%

We evaluate our model in relation to several key metrics - including total accuracy on both positive and negative classes and precision, recall, F1 score and area under the precision recall curve for the positive class. While it is of vital importance to flag as many upcoming personal attacks as possible (recall), it is also of crucial importance to maintain credibility by not flagging innocuous conversations (precision). Thus, we pay particular attention to the area under the precision recall curve (AUPR) as a measure of success. This method is preferable to the ROC curve due to the high class imbalance. The AUPR metric was used heavily in the "Trouble on the Horizon" paper, in which the authors achieved an AUPR of .70 on their final model \cite{chang2019trouble}, although we do not make an explicit comparison to this result due to the disparate nature of our conversational data.

\begin{figure}[htp]
    \centering
    \includegraphics[width=8cm]{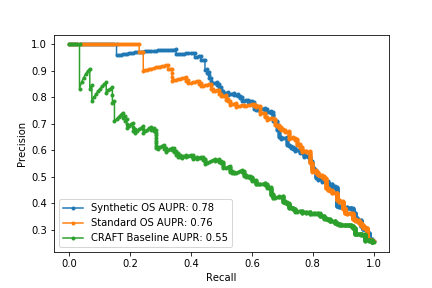}
    \caption{Precision-recall curves and the area under each curve.}
    \label{fig:precision_recall}
\end{figure}

The transformer based BERTweet model outperforms the LSTM based CRAFT model in all metrics. Implementing synthetic oversampling also improves all metrics, with the exception of recall on the positive class on the classification threshold of .5. Given the higher AUPR score, however, we see that synthetic oversampling has generally improved our model's ability to classify the positive class.

\subsection{Analysis}
We now examine the behavior of our model to better understand what aspects of the data it is using for predictions and where it is falling short.

\subsubsection{Model Behavior}
\paragraph{The Lead Up to an Attack}
We first examine whether the model is using the full two tweet context we are supplying to make its predictions or if a single tweet context is sufficient. Intuitively, we expected only a minor drop in performance since our assumption was that most of the signal indicating an attack was coming would be captured in the tweet immediately preceding the attack tweet. We tested this hypothesis by truncating the training data to a single tweet. We were surprised to see a dramatic drop in performance, with the model performing at .74 accuracy and only .50 AUPR. This indicates that the tweet two prior does indeed provide a large amount of useful information in making predictions.

\paragraph{Conversational Dynamics}
Secondly, we look at whether the model is using information about conversational dynamics in the context to inform its predictions. To look at this, we remove the tweet separator token, $</s>$, so that any indication of delineation between speakers is lost. Model performance again drops, although not as significantly as in our first test, with an accuracy of .77 and AUPR of .76. This meets our expectation since the location of inflammatory language that leads to a personal attack  relative to the token (in other words, which tweet in the context it is in) would presumably be valuable information for the model to use in making a prediction on the current tweet. 

\subsubsection{Model Limitations}
\paragraph{Limited Dataset}
While building our model, one of the major challenges we faced was the relatively small size of the dataset we were fine-tuning on, particularly in the positive class. The main consequence of this was a tendency of our model to overfit the training dataset. We dealt with this issue through a variety of methods, including synthetic oversampling (as mentioned earlier), reducing batch size to 10, maintaining the learning rate at the default level of 5e-5 and ensuring that training did not exceed 4 epochs.

Even so, we occasionally see unusual spikes in the negative log likelihood loss pattern on the validation set related to our limited dataset. These spikes would occur despite an increasing accuracy and AUPR score. In other words, our model is making more correct predictions on the positive and negative class on our validation data but is also more over-confident in the bad predictions it is making. We posit two potential explanation for this. Firstly, despite the measures above, our model is likely still overfitting to an extent on the training data. This leads to over-confident incorrect predictions due to random noise as opposed to true signal in our validation data. Secondly, due to the small number of positive examples in our validation set (154), a small change in parameters between epochs could result in dramatic swings in total loss. In other words, a few very bad predictions could have a large impact on the loss calculation, resulting in the spike we witnessed.

\paragraph{GIF Confusion} We observe a higher occurrence of url strings in the context strings where our model makes misclassifications on the test set. The ratio of url strings to context strings is .65 in the misclassified examples vs only .5 in the correctly classified examples. This makes sense intuitively since these url strings contain a large amount of information relative to whether a personal attack is coming that is not interpretable by our model. The best example of this is GIFs, or animated images. While these GIFs are represented in our data as simple url strings, the actual content of the GIF could be highly inflammatory, benign or even friendly, which could be highly indicative of the presence or absence of a personal attack in the subsequent tweet. While we did not have time to explore this further, we believe this would be a fruitful area of future work, as we will discuss in more detail in the future work section.

\paragraph{Nuanced Langugage} In their paper, Price et al. note the difficulty for neural models to identify nuanced language that indicates negative interactions such as sarcasm, dismissiveness, and condescension. {\cite{price2020attributes}}. Notably, BERT was shown by the authors to be particularly poor at identifying sarcasm. These are language qualities that could be highly indicative of impending personal attacks. Since this is not currently accounted for in our model, we believe this is likely a source of confusion contributing to our existing loss.

\section{Conclusions and Future Work}
In summary, we introduced a transformer based model for forecasting conversational events on a novel dataset of Twitter conversations. This model indicates some ability to understand conversational dynamics between speakers. It fills a void in the existing literature in that it provides state-of-the-art predictive performance on Twitter, a platform that has not been studied in the space of conversational forecasting. 

Given Twitter's stated goal of having healthier conversations on its platform, we hope this study is a foundation for future work specific to this ubiquitous channel of communications. One vital area of expansion will be to incorporate additional topics of conversation into the model. While we focused our study on controversial political and societal topics, additional work is needed to ensure the model generalizes to more mundane topics, since the signals leading towards a personal attack could be very different for these conversations.

While our model performed impressively in regards to our test data, we are confident that future work could further improve performance of the model. Our model could be improved by addressing all the sources of loss mentioned earlier. Giving the model a sense of nuanced language by appending an attention head for detecting the 6 attributes of unhealthy conversations as noted in Price et al. would do well to address this deficiency. Another clear area of improvement, as referenced earlier, would be to communicate the sentiment of GIFs as input to the model. Finally, the robustness of the model would be greatly improved by anyone with the resources to reliably collect and accurately tag additional tweets.

As the model becomes more capable, we hope it can become a practical tool to assist Twitter users to interact with more civility and awareness. Given this fine-tuned ability to recognize conversations headed for derailment, Twitter could proactively warn users. We believe that this awareness would allow users to guide themselves towards a more productive resolution, allowing all parties involved to have a better, more positive experience.

\begin{figure}[htp]
    \includegraphics[width=8cm]{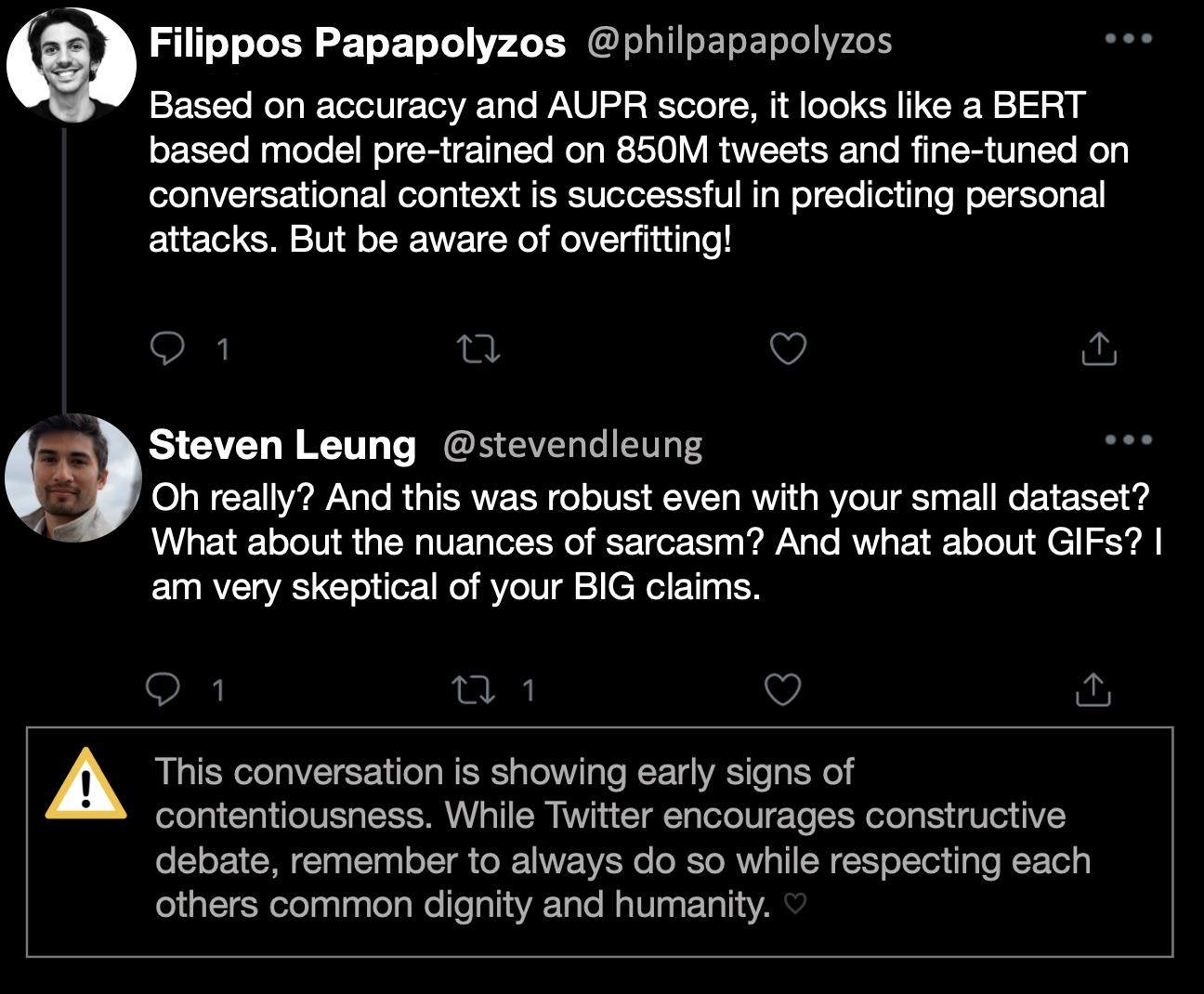}
    \caption{A mock-up of a conversation warning notification on a sample conversation.}
    \label{fig:precision_recall}
\end{figure}

Github repo: \url{https://github.com/stevendleung/Hashing-It-Out/tree/main}

\section{References}

\bibliography{library}

\bibliographystyle{abbrv}

\end{document}